\documentclass[11pt, a4paper, logo, copyright, nonumbering]{deepseek}
\usepackage[authoryear, sort&compress, round]{natbib}
\usepackage{dblfloatfix}
\usepackage{ulem}
\usepackage{caption}
\usepackage{dramatist}
\usepackage{xspace}
\usepackage{pifont}
\usepackage{multirow}
\usepackage{adjustbox}
\usepackage{tcolorbox}
\usepackage{xltabular}
\usepackage{longtable}
\usepackage{makecell}
\usepackage{algorithm}
\usepackage[noend]{algpseudocode}

\usepackage{amsfonts}
\usepackage{amsmath}
\usepackage{amssymb}
\usepackage{lineno}

\usepackage[bottom]{footmisc}

\usepackage{CJKutf8}
\usepackage{subfigure}
\usepackage{setspace}

\usepackage{titlesec}
\usepackage{enumitem}

\usepackage[colorlinks,linkcolor=orange,anchorcolor=blue,citecolor=lightgray]{hyperref}

\makeatletter
\def\@BTrule[#1]{%
  \ifx\longtable\undefined
    \let\@BTswitch\@BTnormal
  \else\ifx\hline\LT@hline
    \nobreak
    \let\@BTswitch\@BLTrule
  \else
     \let\@BTswitch\@BTnormal
  \fi\fi
  \global\@thisrulewidth=#1\relax
  \ifnum\@thisruleclass=\tw@\vskip\@aboverulesep\else
  \ifnum\@lastruleclass=\z@\vskip\@aboverulesep\else
  \ifnum\@lastruleclass=\@ne\vskip\doublerulesep\fi\fi\fi
  \@BTswitch}
\makeatother

\addto\extrasenglish{
}

 {\begin{list}{}%
         {\setlength{\leftmargin}{#1}}%
         \item[]%
 }
 {\end{list}}

\reportnumber{001} 

\renewcommand{\today}{}
\newcommand{\spmath}{OpusAnimation}

\title{\centering \spmath: Code-Based Dynamic Chart Generation}
\vspace{-5mm}
\author[*]{
\small
\hspace{2em}
Bozheng Li$^{1,2}$, Miao Yang$^{1}$, Zhenhan Chen$^{1}$, Jiawang Cao$^{1}$, Mushui Liu$^{3}$
\newline
Yi Lu$^{1,4}$,
Yongliang Wu$^{1}$,
Bin Zhang$^{1}$,
Yangguang Ji$^{1}$, 
Licheng Tang$^{1}$,
Jay Wu$^{1}$,
Wenbo Zhu$^{1}$

\small
$^1$Opus AI Research, $^2$Brown University, $^3$Zhejiang University $^4$ University of Toronto \\
\small
\texttt{\{bozheng\_li\}@brown.edu} \\
\small
}

\renewcommand{\phi}{\varphi}

\renewcommand{\epsilon}{\varepsilon}
\renewcommand{\imath}{\mathrm{i}}

\newlength{\restsubwidth}
\newlength{\restsubheight}
\newlength{\restsubmoreheight}
\newcommand{\rest}[2]{%
        \settowidth{\restsubwidth}{\ensuremath{#2}}
        \settoheight{\restsubheight}{\ensuremath{{}_{#2}}}
        \ensuremath{{#1\hskip 0.5pt}_{\vrule\kern2pt\parbox[b][%
        4pt][b]{\the\restsubwidth}{%
                        \ensuremath{{}_{#2}}}}}
        }

\begin{abstract}
Dynamic Chart Generation (DCG) involves producing code-rendered animated visualizations as charts. While recent advances in multi-modal large language models (MLLMs) have significantly improved their capability on static chart generation and comprehension, MLLMs' potential for handling dynamic chart generation and understanding remains underexplored. To bridge this research gap, we introduce \textbf{DCG-Bench} (\textbf{D}ynamic \textbf{C}hart \textbf{G}eneration \textbf{Bench}mark), the first benchmark evaluating MLLM's capability on dynamic chart generation tasks from three dimensions: \textit{Simple Text-to-Chart}, \textit{Detailed Text-to-Chart}, and \textit{Video-to-Chart} tasks. We construct DCG-8K, a high-quality DCG dataset with annotations covering instruction-code-video triplets and QA pairs for both code and video evaluation. Based on DCG-8K, we explored a two-stage training recipe, proposing Joint-Code-Visual Reward for group relative policy optimization to construct expert MLLM Qwen2.5-VL-DCG-3B for the DCG task. Our benchmarking result reveals shortcomings of existing MLLMs in the visual-to-chart task, and our model beats the best open-sourced MLLM with an average 8.31\% performance gain across three tasks, and shows on par performance against proprietary models with only 3B parameters, proving the effectiveness of our training recipe.  Our code and dataset will be publicly available.
 \end{abstract}

\begin{document}
\begin{CJK*}{UTF8}{gbsn}

\maketitle

\begin{figure*}[h]
    \centering
    \includegraphics[width=0.85\textwidth]{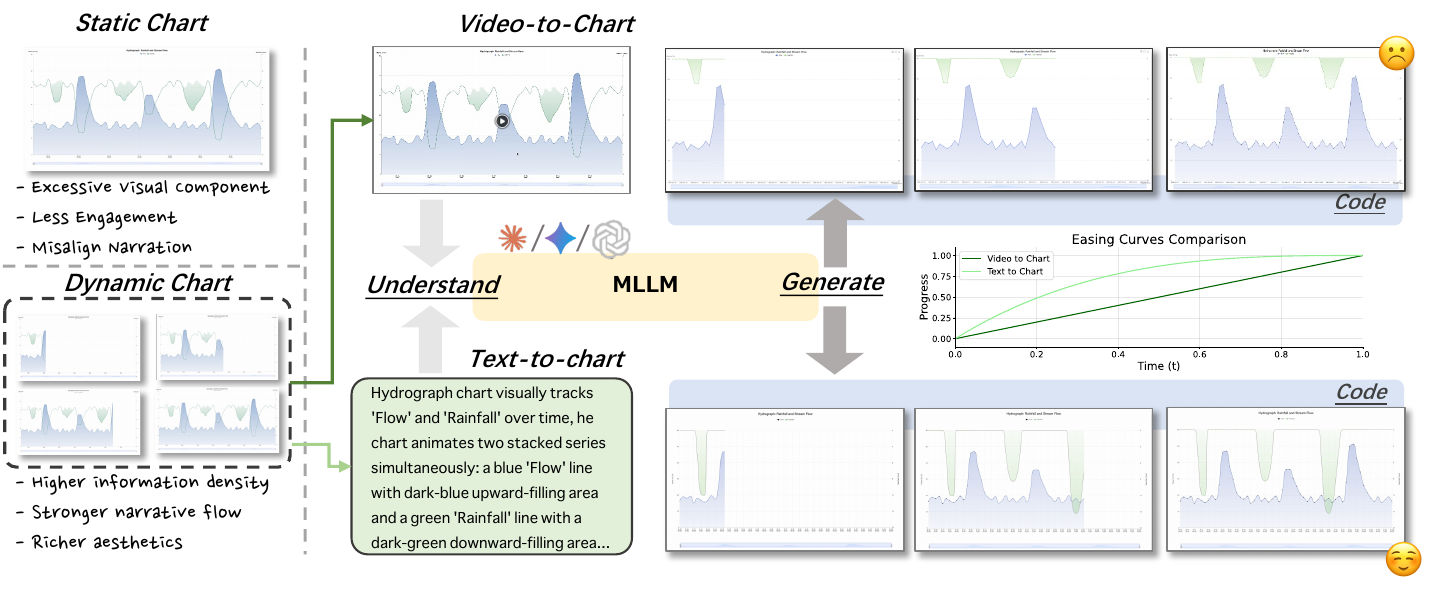}
    \caption{
        Illustration of the Dynamic Chart Generation Task, and the bad case of existing MLLM on the V2C task.
    }
    \label{fig:teaser}
\end{figure*}
\newpage


\newpage

\section{Introduction}

Charts are an essential visualization tool for effective information communication. Significant progress has been made in understanding~\cite{masry2023unichart, kantharaj2022chart, han2023chartllama} and generating~\cite{zhao2025chartcoder, xia2024chartx, yang2024chartmimic} static charts, with recent Multimodal Large Language Models (MLLMs)\cite{zhao2025chartcoder, xia2024chartx} demonstrating strong capabilities in this area. Beyond static charts, dynamic charts convey higher information density\cite{amini2018hooked}, provide stronger narrative flow~\cite{srinivasan2018augmenting}, and deliver richer aesthetics~\cite{ren2017chartaccent} through animation, significantly boosting user attention and memory retention~\cite{chevalier2016animations} across practical scenarios such as video production, interactive web pages, and presentation systems. However, the code-based dynamic chart generation task remains under-explored.

To fill this gap and push boundaries of MLLM capabilities in dynamic chart generation, we introduce \textbf{D}ynamic \textbf{C}hart \textbf{G}eneration Benchmark (\textbf{DCG-Bench}) to evaluate the performance of MLLM on the DCG task. DCG task comprises three categories: \textit{Detailed text-to-Chart (D2C)}, \textit{Simple text-to-Chart (S2C)} and \textit{Video-to-Chart (V2C)}, covering different granularity and modality of input instruction. DCG-Bench samples contain instruction-code-video triplets with both detailed and simple instructions, accompanied by 10 questions for code and 10 questions for video, verifying whether generated results meet requirements, enabling evaluation of MLLM generation quality through QA metrics from both perspectives.

For benchmark construction, data acquisition is challenging. Crawling well-organized data from the internet is difficult, while generating dynamic charts from scratch using proprietary models~\cite{GPT4o, Claude3.7, team2023gemini} remains unstable, as demonstrated by the zero-shot results in Table~\ref{tab:Main_Result}. Inspired by previous work~\cite{zhao2025chartcoder}, we developed a four-stage pipeline: sourcing and cleaning chart templates, applying random animation modifications, rendering code into videos while extracting paired descriptions, and generating QA pairs for evaluation. Through this pipeline, we expanded a small number of well-defined seed templates~\cite{Echart} into DCG-8K, a well-crafted DCG dataset with 8K high-quality and diversified samples, of which 700 comprise DCG-Bench. As shown in Table~\ref{tab:Main_Result} and Figure~\ref{fig:teaser}. Our preliminary evaluations on DCG-Bench reveal that while commercial models~\cite{GPT4o, Claude3.7} demonstrate strong performance on textual-instruct generation, they exhibit weaker results on visual-based generation. This observation motivates us to explore a training recipe for developing expert MLLMs that can generate code-based dynamic charts from both textual and visual instructions.

Since the DCG task requires generating complex HTML code with outputs often exceeding thousands of tokens, relying solely on supervised fine-tuning to build an expert MLLM on DCG data risks overfitting. This is due to memorization artifacts~\cite{perez2023discovering, ouyang2022training}, where the model tends to remember answers content rather than learning task-relevant patterns, especially when training samples are long or overly detailed. Therefore, we explore a two-stage training recipe. We first align the base MLLM to the DCG task through supervised fine-tuning (SFT), then in the second stage we propose a multi-modal Joint-Code-Visual Reward, which evaluates the generated dynamic chart's quality from both code and visual perspectives using comprehensive QA pairs to apply Group Relative Policy Optimization (GRPO) that further enhances the performance and generalizability of the model. Leveraging our training recipe and multi-modal reward, we constructed Qwen2.5-VL-DCG-3B, which demonstrates competitive performance against larger MLLM~\cite{bai2025qwen2} on V2C task, as well as CodeLLM~\cite{hui2024qwen2} on D2C and S2C tasks, achieving an average 8.31\% performance gain across three tasks while maintaining a lightweight model footprint. Our experiments yield several important insights for DCG training. First, incorporating training data from multiple modalities enhances the model's base DCG capabilities in cold start stage. Second, GRPO training demonstrates strong generalization ability, effectively transferring generation skills learned from textual instructions to video-based queries. Third, introducing code-level reward as a process-level signal proves particularly beneficial under GRPO training, while a video-based reward further ensures the overall quality of the generated outputs.

\noindent In summary, the main contributions of this work are as follows:
\begin{itemize}
    \item We introduce DCG-Bench, the first benchmark designed to evaluate the performance of MLLM on the dynamic chart generation task, covering instruction type across detail text-to-chart, simple text-to-chart, and video-to-chart tasks, along with an 8K high-quality dataset to advance research in dynamic chart generation.
    \item We propose a novel Joint-Code-Visual Reward enabling the interaction between Code Quality and Visual Quality Assessment, exploring the potential of reinforcement learning of MLLM on an integrated application scenario like dynamic chart generation.
    \item We provide a detailed training recipe to construct an expert MLLM for dynamic chart generation, achieving competitive performance to 32B models with only 3B parameters.
\end{itemize}

\section{Related Work}
\label{sec: related_work}

\noindent
\textbf{Code-based Chart Generation.} As a vital form of data visualization, chart-related understanding~\cite{masry2023unichart, kantharaj2022chart} and code-based chart generation~\cite{zhao2025chartcoder, xia2024chartx, yang2024chartmimic, li2025metal, wei2024words, zhang2024chartifytext, galimzyanov2024drawing, yang2024matplotagent} tasks have been extensively explored in prior works. For the generation task, ChartMimic~\cite{yang2024chartmimic} proposes a benchmark to assess the chart-to-code generation capabilities of MLLMs, while ChartCoder~\cite{zhao2025chartcoder}, ChartVLM~\cite{xia2024chartx}, and META~\cite{li2025metal} introduce specifically designed MLLMs or agentic systems to produce chart-rendering code. However, these methods primarily target static chart, overlooking the growing importance of dynamic charts in real-world applications~\cite{wang2024wonderflow, ren2017chartaccent}. The data visualization community has attempted to support dynamic chart generation through pipeline-based systems~\cite{ying2024reviving, ku2025theoremexplainagent}, but the capability of MLLMs to directly generate dynamic chart code from scratch remains underexplored. 

\noindent
\textbf{Multimodal Code Benchmarks.} Recently, multimodal coding benchmarks~\cite{li2024mmcode, zhang2024humaneval, yang2024swe} have been proposed to evaluate the ability of MLLMs to solve coding problems that include visual inputs. Among them, Design2Code~\cite{si2024design2code} and Web2Code~\cite{yun2024web2code} focus on HTML generation from UI designs, while ChartMimic~\cite{yang2024chartmimic}, ChartX~\cite{xia2024chartx}, and Plot2Code~\cite{wu2024plot2code} evaluate chart-to-code generation using Python. Our work extends this line by introducing a benchmark for dynamic chart generation, which evaluates MLLMs' ability to generate JavaScript-enhanced HTML code from multimodal input. 
The DCG task requires capturing temporal dynamics. For instance, a growing line chart with elastic easing reflects time-dependent motion that requires fine-grained video understanding to reproduce.

\noindent
\textbf{Group Relative Policy Optimization.} GRPO~\cite{shao2024deepseekmath} simplifies PPO~\cite{schulman2017proximal} by replacing the value function with the average reward from policy rollouts as the baseline for advantage estimation. Recent studies~\cite{zhang2025r1, wang2025vl, li2025videochat, liao2025improved} have applied GRPO to base vision tasks like ocr or object counting. GRPO exhibit high generalizatiobn ability~\cite{chu2025sft}, as well as high data-efficient training using limited data~\cite{wang2025reinforcement}. These characteristics are well-suited for DCG tasks, which involve long output sequences and are prone to overfitting and memorization artifacts~\cite{perez2023discovering, ouyang2022training}.

\section{Dataset and Benchmark}
\label{sec: dataset_and_benchmark}

\begin{figure*}[t]
    \centering
    \includegraphics[width=\textwidth]{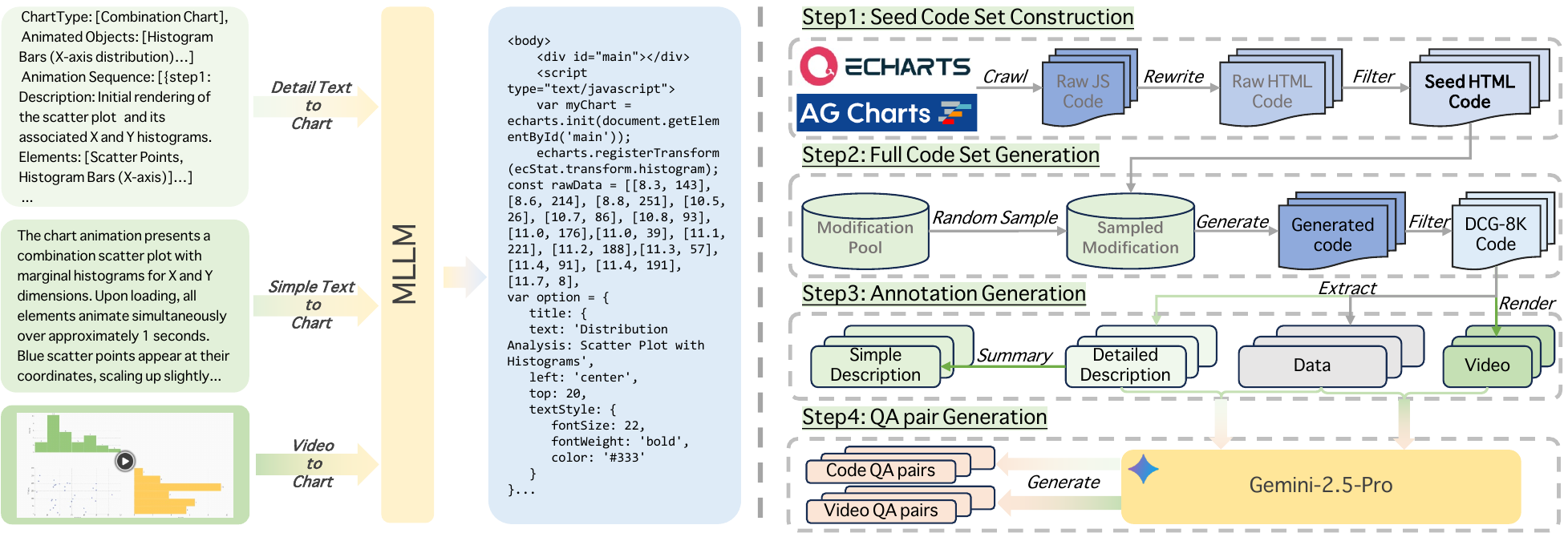}
    \caption{
        Demonstration of three DCG task types and our data curation pipeline. \textbf{Left:} Three task types (Detail Text to Chart, Simple Text to Chart, and Video to Chart) in DCG-Bench. \textbf{Right:} Dataset curation pipeline from raw ECharts code to final DCG-8K dataset.
    }
    \label{fig:dataset_pipeline}
\end{figure*}

\subsection{Overview}
DCG-8K is a comprehensive dataset curated for dynamic chart generation tasks. Each sample in the DCG-8K dataset is composed of: an HTML code snippet $c$, the corresponding rendered animation video $v$, detailed text descriptions $t_d$, simple text descriptions $t_s$, question-answer pairs $QA_{code}$ for code evaluation, and $QA_{video}$ for video evaluation. The test split of DCG-8K functions as DCG-Bench for evaluating the performance of MLLM on DCG tasks.

\subsection{Task Definition}
As shown in Figure~\ref{fig:dataset_pipeline}, DCG-Bench comprises three tasks: detail text-to-chart, simple text-to-chart, and video-to-chart. For each task, the model receives a query $q$ consisting of a data sequence $d$ and descriptions that vary in modality and granularity. The model is expected to generate complete HTML code $c_g$ that renders a video animation $v_g$ displaying a dynamic chart visualization of $d$ while matching the provided description.

\noindent
\textbf{Detailed Text-to-Chart (D2C).} Description for D2C task is detailed textual description $t_d$ specifying chart type, animated objects, size modifications, color transitions, shape details and animation effects. D2C task evaluates the capability of the model in following specific generation requirements.

\noindent
\textbf{Simple Text-to-Chart (S2C).} Description for S2C task is simple description $t_s$ summarized from $t_d$ in D2C task. By replacing $t_d$ with $t_s$, S2C task mimics realistic human-level instructions, measuring model's capability in parsing simple instructions into high-quality dynamic chart animations.

\noindent
\textbf{Video-to-Chart (V2C).} The description for V2C task is a reference video $v$ displaying dynaimc chart visualization of $d$. V2C task requires MLLMs to capture visual information, including motion and animation effects, to reproduce the provided dynamic chart, assessing the video understanding and code generation capabilities of the model.

\subsection{Metrics}
\label{sec:metrics}
To evaluate quality of generated code $c_g$ and rendered chart animations $v_r$, two metrics are employed:

\noindent
\textbf{Execution Pass Rate.} We first assess basic executability by verifying whether generated code $c_g$ renders a valid, non-blank animation video $v_g$. Each instance then receives a binary indicator, and overall pass rate is computed as proportion of successful executions across the test samples.

\noindent
\textbf{QA-based Scores.} To better assess the semantic alignment between generated outputs and the input specifications in dynamic chart generation, we propose a QA-based metric using powerful MLLM~\cite{team2023gemini}. For each chart, we construct around ten targeted QA pairs based on its reference code or video, focusing on fine-grained aspects such as animation order, element appearance, and timing consistency \textit{(e.g., “Do the axes and gridlines appear before the histogram bars begin animating?”)}. Each question is answered by the MLLM using the generated output, and a binary score is assigned (1 for correct, 0 otherwise). This approach has previously been adopted by various generation tasks~\cite{chai2024auroracap, wu2024towards}, providing a more interpretable and task-specific evaluation than traditional rule-based GPT scoring~\cite{xia2024chartx, yang2024chartmimic} or feature-similarity metrics like FVD~\cite{unterthiner2019fvd} and CLIPSim~\cite{wu2021godiva}, which are less effective for chart animation due to limited domain alignment. Chart generation inherently involves complex visual or textual inputs that map to relatively deterministic visual behaviors, making decomposition-based QA evaluation a more suitable and objective choice~\cite{chai2024auroracap, wu2024towards, team2023gemini}.

Formally, we derive two sets of QA pairs from the input query $q$: $N_c$ code-based QA pairs $QA_{\text{code}}$ targeting the generated code $c_g$, and $N_v$ video-based QA pairs $QA_{\text{video}}$ targeting the generated video $v_g$.
Based on these QA sets, we define the evaluation scores $S_{\text{code}}$ and $S_{\text{video}}$ as:
\begin{align}
    S_{code}(q, c_g) &= \frac{1}{N_{c}}\sum^{N_c}_{i=1} \text{MLLM}_{eval}(c_g, QA^{(i)}_{code}) \label{eq:code_score}  \\
    S_{video}(q, v_g) &= \frac{1}{N_{v}}\sum^{N_v}_{j=1} \text{MLLM}_{eval}(v_g, QA^{(j)}_{video}) \label{eq:video_score}
\end{align}
Here, $\text{MLLM}_{eval}$ employs Gemini2.5-Pro~\cite{team2023gemini} and returns 1 if the artifact meets the corresponding QA requirement, and 0 otherwise. Specifically, if the $c$ failed to render, both scores will be 0.

\subsection{Dataset Construction}
As shown in Figure~\ref{fig:dataset_pipeline}, the dataset construction pipeline involves four main steps. \textbf{Firstly}, We crawled a set of JavaScript chart templates from ECharts~\cite{Echart}, manually cleaned and converted them into HTML snippets $C_{\text{seed}}$, and filtered them for complexity and diversity. This resulted in 175 seed templates spanning 18 chart categories. \textbf{Secondly}, we defined a pool of animation modifications $m$ covering various aspects including \textit{data elements}, \textit{layout}, \textit{visual style}, \textit{animation speed}, and \textit{animation effects}. For each $c_{\text{seed}}$, we randomly sampled one to ten modifications prompted Gemini-2.5-Pro~\cite{team2023gemini} to apply them, generating 8,000 modified HTML code samples $C$. Non-rendering samples were filtered out. \textbf{Thirdly}, each valid code sample $c$ was rendered into a dynamic chart animation video $v$ using a headless browser with Timesnap~\cite{timesnap} and FFmpeg~\cite{FFmpeg}. Gemini-2.5-Pro then generated a detailed textual description $t_d$, which is summarized into simple descriptions $t_s$, and extracted the data sequence $d$ used in the chart animation. \textbf{Finally}, based on the code-video pair $(c, v)$, we generated 10 question–answer pairs for both modalities to assess whether a generated result matches the provided requirements. DCG-8K covers 20 chart types, with reference code lengths averaging over 2000 tokens, highlighting diversity and complexity of DCG-8K. Further details on tool use and prompt sets, and detailed analysis of dataset statistics is included in Appendix.

\section{Training Recipe}
\label{sec: training_recipe}
\begin{figure*}[t]
    \centering
    \includegraphics[width=0.95\textwidth]{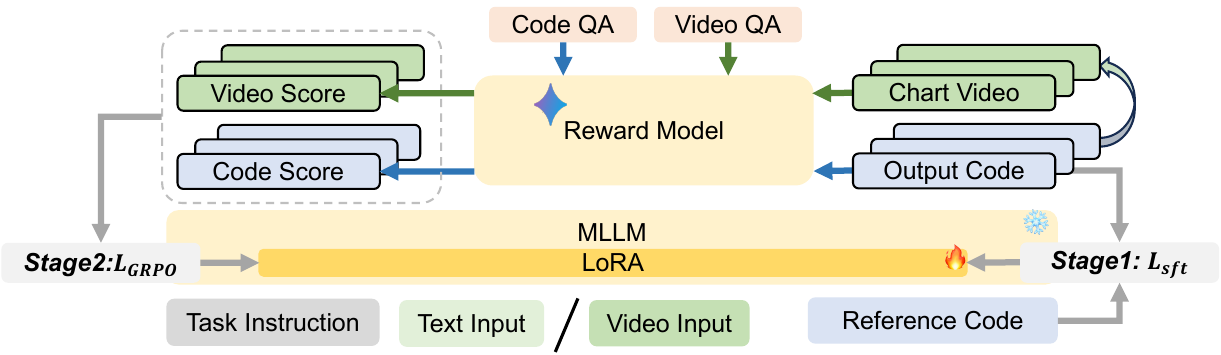} 
    \caption{
        Training Recipe of Qwen2.5-VL-DCG, including SFT and JCV-GRPO training
    }
    \label{fig:method}
\end{figure*}

Based on DCG-8K, we explored a two-stage training recipe to construct expert MLLM for DCG task based on Qwen2.5-VL-3B~\cite{bai2025qwen2}. A shown in Figure~\ref{fig:method}, we first perform supervised fine-tuning (SFT) on D2C and V2C tasks, providing the base model with a cold-start initialization for DCG tasks. Then we apply Joint-Code-Visual Reward based Group Relative Policy Optimization (JCVR-GRPO) to further enhance performance and generalization capability of model. We refer to the models obtained after SFT and JCVR-GRPO stages as Qwen2.5-DCG-3B$_{s1}$ and Qwen2.5-DCG-3B$_{s2}$, respectively.

\subsection{Code Generation Supervised Fine-tuning}
In the first stage, we apply SFT to enable the model to generate renderable code while better following task instructions from both textual and visual modalities. The training data comprises 8K instances derived from 4K samples across D2C and V2C tasks. S2C task is excluded to simulate real-world prompting scenarios and evaluate the generalization ability of the model, while D2C and V2C provide richer supervision for learning accurate instruction-to-code mappings.

\subsection{Joint-Code-Visual Reward based Group Relative Policy Optimization}
\textbf{GRPO~\cite{guo2025deepseek}} simplifies PPO~\cite{schulman2017proximal} by replacing the value function with the average reward of policy samples as the baseline for computing advantages. Formally, when presented with a query $q$, the method generates multiple response samples $\{o_1, o_2, \cdots, o_G\}$ and evaluates their corresponding rewards $\{r_1, r_2, \cdots, r_G\}$ using a reward model. The advantage is calculated through normalization: $\hat{A}_{i,t} = \frac{r_{jcv}^i - \text{mean}(\mathbf{r_{jcv}})}{\text{std}(\mathbf{r_{jcv}})} \label{eq:advantage} $. The model weight is optimized using fthe ollowing objective~\cite{shao2024deepseekmath}:
\begin{gather}
\mathcal{J}_{\text{GRPO}}(\theta) = \mathbb{E}_{q\sim P(q),\{o_i\}_{i=1}^G \sim \pi_{\text{old}}(O|q)} \left[ \frac{1}{G}\sum_{i=1}^G \frac{1}{|o_i|}\sum_{t=1}^{|o_i|}\left\{\min\left[\frac{\pi_\theta(o_{i,t}|q,o_{i,<t})}{\pi_{\text{old}}(o_{i,t}|q,o_{i,<t})}\hat{A}_{i,t},\right.\right.\right. \notag \\
\left.\left.\left.\text{clip}\left(\frac{\pi_\theta(o_{i,t}|q,o_{i,<t})}{\pi_{\text{old}}(o_{i,t}|q,o_{i,<t})},1-\epsilon,1+\epsilon\right)\hat{A}_{i,t}\right]-\beta\mathbb{D}_{KL}\left[\pi_\theta\|\pi_{\text{ref}}\right]\right\}\right]
\end{gather}
where $\pi_\theta$ and $\pi_{\text{old}}$ are current and old policy, $\epsilon$ and $\beta$ are hyper-parameters introduced in PPO.

\noindent
\textbf{Joint-Code-Visual Reward.} DCG tasks inherently involve long output sequences with complex code details, which correspond to a larger target generation space, increasing the likelihood of memorization artifacts~\cite{perez2023discovering, ouyang2022training}. In contrast, GRPO employ goal-conditioned optimization rather than imitating a single reference output, thereby exhibiting superior generalization ability compared to SFT\cite{shao2024deepseekmath, chu2025sft}. To mitigate the overfitting issues, we adopt GRPO for further training. For DCG task, we specifically design the Joint-Code-Visual Reward $r_{jcv}(q, c, v)$ which jointly evaluates the functional correctness of the generated code $c_g$ and the visual fidelity of the rendered video $v_g$ with respect to the input query $q$. JCVR apply Gemini2.5-Pro~\cite{team2023gemini} as reward model and integrate two quality scores introduced in Section~\ref{sec:metrics}:
\begin{align}
    r_{jcv}(q, c, v) = w_{code} \cdot s_{code}(q, c) + w_{video} \cdot s_{video}(q, v) \label{eq:jcv_reward}
\end{align}
Here, $w_{code}$ and $w_{video}$ are coefficients balancing the code and visual reward signal. When generated code fails to render video, both scores are set to $-0.25$, penalizing non-executable code. JCV reward provides both process-level and output-level supervision for DCG tasks. $S_{\text{code}}$ reflects the quality of the generation process and functions as a process reward, while $S_{\text{video}}$ evaluates the final rendered output and serves as an output reward. By combining these two components, JCV delivers a dense and informative reward signal that supports generalizable training. Formal proof of the equivalence between JCV and the combination of PRM and ORM is provided in the appendix. JCV-GRPO training is conducted on rest 1K D2C samples from 5K training split of DCG-8K.

\section{Experiment}
\label{sec:experiments}

\subsection{Baseline and Dataset}
Extensive experiments are conducted to benchmark baseline model performance across three DCG tasks and assess the effectiveness of our training recipe via DCG-Bench. Baseline models encompass proprietary models include Gemini~\cite{team2023gemini}, GPT~\cite{GPT4o}, and Claude~\cite{Claude3.7}, along with open-source models like CodeLLM Qwen-2.5-coder~\cite{hui2024qwen2}, as well as MLLM including Qwen-VL2.5~\cite{bai2025qwen2} and InternVL3~\cite{zhu2025internvl3}. For Qwen2.5-DCG-3B, we perform zero-shot evaluation on the S2C task, while the other two tasks are included in training data. The DCG-8K dataset is split into 5K training samples (4k for SFT and 1K for GRPO), 2.3K validation samples, and 700 test samples, with test set forming DCG-Bench.

\subsection{Implementation Details}
We employ Qwen2.5-VL-3B~\cite{bai2025qwen2} as base MLLM and apply LoRA~\cite{hu2022lora} to update the LLM parameters while keeping the vision encoder and adapter frozen. For SFT, we apply LoRA with $r=64$ and $\alpha=128$, with learning rate of $1e^{-4}$. For JCV-GRPO training, we apply LoRA with $r=32$ and $\alpha=64$, with learning rate of $1 \times 10^{-5}$, group size of 8, $\beta=0.04$ and $\epsilon=0.2$ with max output length set to 8192 to guarantee diverse and complete output. Animation Videos are rendered at 24 FPS with a 2-second duration. We leverage Gemini-2.5-Pro-3-25-preview~\cite{team2023gemini} as reward model to generate code and video scores. More implementation details are provided in the appendix.

\subsection{Main Experiment}
\definecolor{bestHighlightColor}{HTML}{FDCBCB} 
\definecolor{secondHighlightColor}{HTML}{FFE5CC} 

\begin{table}
    \centering
    \begin{adjustbox}{width=\textwidth}
    \begin{tabular}{l|ccc|ccc|ccc}
        \toprule
        \multirow{2}{*}{Method} & \multicolumn{3}{c|}{D2C} & \multicolumn{3}{c|}{S2C} & \multicolumn{3}{c}{V2C} \\
        & $Eexc. Rate $ & $S_{code}$ & $S_{video}$ & $Eexc. Rate $ & $S_{code}$ & $S_{video}$ & $Eexc. Rate $  & $S_{code}$ & $S_{video}$ \\
        \midrule
        Full Score                             & 100 & 10 & 10 & 100 & 10 & 10 & 100 & 10 & 10 \\
        \midrule
        \rowcolor{gray!20}\multicolumn{10}{c}{Proprietary} \\
        \midrule
        GPT4.1~\cite{GPT4o}                    & 94.40 & 8.20 & 5.67 & 93.10 & 5.93 & 4.81 & 96.41 & 3.01 & 4.98 \\
        Claude~\cite{Claude3.7}               & 94.97 & 8.39 & 6.07 & 92.53 & 6.28 & 5.64 & 94.53 & 3.06 & 4.73 \\
        Gemini-2.5-flash~\cite{team2023gemini}             & 81.47 & 7.62 & 6.13 & 76.14 & 5.80 & 5.28 & 75.14 & 2.66 & 4.68 \\
        Gemini-2.5-pro~\cite{team2023gemini}               & 91.95 & 8.71 & 6.85 & 87.64 & 6.75 & 6.01 & 85.92 & 3.35 & 6.06 \\
        \midrule
        \rowcolor{gray!20}\multicolumn{10}{c}{CodeLLM} \\
        \midrule
        Qwen2.5-coder-32B~\cite{hui2024qwen2}           & 88.97 & 7.23 & 6.32 & 90.66 & 6.07 & 6.21 & - & -& -\\
        Qwen2.5-coder-14B~\cite{hui2024qwen2}           & 89.37 & 6.51 & 6.26 & 88.65 & 5.67 & 6.07  & - & -& -\\
        Qwen2.5-coder-7B~\cite{hui2024qwen2}           & 83.62 & 4.63 & 5.35 & 86.21 & 4.97 & 5.57  & - & -& -\\
        Qwen2.5-coder-3B~\cite{hui2024qwen2}            & 73.99 & 3.51 & 4.66 & 69.4 & 3.27 & 4.20  & - & -& -\\
        \midrule
        \rowcolor{gray!20}\multicolumn{10}{c}{MLLM} \\
        \midrule
        InternVL3-14B~\cite{zhu2025internvl3}            & 66.52 & 3.35 & 3.64 & 65.23 & 3.34 & 3.67 & 69.11 & \cellcolor{orange!30}3.52 & 3.98 \\
        InternVL3-8B~\cite{zhu2025internvl3}             & 63.94 & 2.79 & 3.55 & 63.36 & 2.82 & 3.53 & 64.22 & 2.88 & 3.44 \\
        InternVL3-2B~\cite{zhu2025internvl3}             & 43.82 & 1.43 & 2.52 & 42.24 & 1.38 & 2.49 & 49.35 & 1.56 & 2.77 \\
        Qwen2.5-VL-32B~\cite{bai2025qwen2}              & \cellcolor{orange!30}88.22 & \cellcolor{orange!30}6.99 & \cellcolor{orange!30}6.25 & \cellcolor{red!50}89.51 & \cellcolor{orange!30}5.70 & \cellcolor{red!50}5.98 & \cellcolor{red!50}94.11 & 3.19 & \cellcolor{yellow!30}5.65 \\
        Qwen2.5-VL-7B~\cite{bai2025qwen2}               & 66.67 & 3.08 & 4.07 & 60.60 & 2.73 & 3.92 & 85.34 & 2.32 & 4.97 \\
        Qwen2.5-VL-3B~\cite{bai2025qwen2}               & 8.91 & 0.37 & 0.50 & 2.59 & 0.22 & 0.15 & 71.12 & 1.58 & 3.76 \\
        \midrule
        \textbf{Qwen2.5-DCG$_{s1}$}        & \cellcolor{yellow!30}87.36 & \cellcolor{yellow!30}6.73 & \cellcolor{yellow!30}5.78 & \cellcolor{yellow!30}82.76 & \cellcolor{yellow!30}5.03 & \cellcolor{yellow!30}4.82 & \cellcolor{yellow!30}87.78 & \cellcolor{yellow!30}3.82 & \cellcolor{orange!30}4.98 \\
        \textbf{Qwen2.5-DCG$_{s2}$}     & \cellcolor{red!50}91.95 & \cellcolor{red!50}7.45 & \cellcolor{red!50}6.61 & \cellcolor{orange!30}89.37 & \cellcolor{red!50}5.77 & \cellcolor{orange!30}5.78 & \cellcolor{orange!30}92.39 & \cellcolor{red!50}4.32 & \cellcolor{red!50}5.66 \\
        \bottomrule
    \end{tabular}
    \end{adjustbox}
\vspace{5pt}
\caption{Performance comparison across properitary model, CodeLLM adn MLLM on DCG-Bench. \colorbox{red!50}{Red}, \colorbox{orange!30}{Orange} and \colorbox{yellow!30}{Yellow} represent best, second and third performance of MLLM.}
\label{tab:Main_Result}
\end{table}

\subsubsection{Benchmarking Existing Model.}
On DCG-Bench, proprietary models exhibit high execution rates, with GPT-4.1~\cite{GPT4o} leading at an average of 94.6\% across three tasks, while open-source models lag slightly behind. The best open-source model Qwen2.5-Coder-32B~\cite{hui2024qwen2} has an average execution rate 4\% lower than Claude-3.7-Sonnet~\cite{Claude3.7}. On the D2C task, models achieve higher execution rates and generation quality than on the S2C task, which we attribute to the limited ability of current models to parse requirements from condensed instructions. In the V2C task, MLLMs achieve higher execution rates than both D2C and S2C. Notably, Qwen2.5-VL-3B~\cite{bai2025qwen2} shows a gap of over 60\% in execution rate between the V2C and D2C tasks. Commercial models perform well on D2C and S2C tasks, with Gemini-2.5-Pro~\cite{team2023gemini} achieving an average score of 7.73 on $S_{\text{code}}$ and 6.43 on $S_{\text{video}}$. Among open-source models, Qwen2.5-Coder-32B~\cite{hui2024qwen2} performs best on D2C and S2C tasks. In the V2C task, Qwen2.5-VL-32B leads among open-source models and shows performance comparable to proprietary models. However, even the best model, Gemini-2.5-Pro, achieves only 3.35 on $S_{\text{code}}$ in the V2C task, indicating the limited capability of current MLLMs to generate dynamic charts from visual input. Meanwhile, for MLLM, reducing model size leads to a larger performance drop in the D2C and S2C tasks than in the V2C task. We attribute this to the visual input of the V2C task: since visual features remain constant across MLLMs of different LLM sizes, instruction information is comparatively better retained.

\subsubsection{Performance of Qwen2.5-DCG-3B}
\textbf{Detail Text-to-Chart (D2C).} As indicated in Table~\ref{tab:Main_Result}, SFT aligns base MLLM with dynamic chart code generation task, bringing 6.36 and 5.28 improvement on $S_{\text{code}}$ and $S_{\text{video}}$, significantly overcoming the weakness of 3B model on DCG task. JCVR-GRPO further pushes Qwen2.5-DCG-3B to achieve the best performance among open-source models, surpassing Qwen-coder-32B~\cite{hui2024qwen2} by 3.73\% on execution rate, and generating high-quality code with 0.46 higher $S_{code}$ and 0.36 higher $S_{video}$, proving the high quality of DCG-8K dataset as well as JCVR-GRPO training recipe.

\noindent
\textbf{Simple Text-to Chart (S2C).} Qwen2.5-DCG-3B maintains strong performance under zero-shot evaluation setting of S2C task with 89.37\% execution rate, achieving competitive performance with 0.07 higher $S_{code}$ than Qwen2.5-VL-32B, demonstrating better instruction following and effective generalization ability acquired through JCVR-GRPO that brings 6.61\% improvement on execution rate and 0.96 on $S_{\text{video}}$ with GRPO training conducted solely on D2C task.

\noindent
\textbf{Video-to-Chart (V2C).} Although JCVR-GRPO is condutced exclusively on D2C data, Qwen2.5-DCG-3B demonstrates remarkable generalization to V2C task, achieving the highest $S_{\text{code}}$ and $S_{\text{video}}$ scores among open-source models, and surpassing Gemini-2.5-Pro by 1.02 on $S_{\text{code}}$. This improvement underscores the effectiveness of our training recipe in fostering robust dynamic chart generation capabilities that generalize across input modalities, thereby mitigating the challenge MLLMs face in translating visual input into dynamic charts.

\begin{figure*}[htp]
    \centering
    \begin{minipage}[b]{0.31\textwidth}
        \centering
        \includegraphics[width=\textwidth]{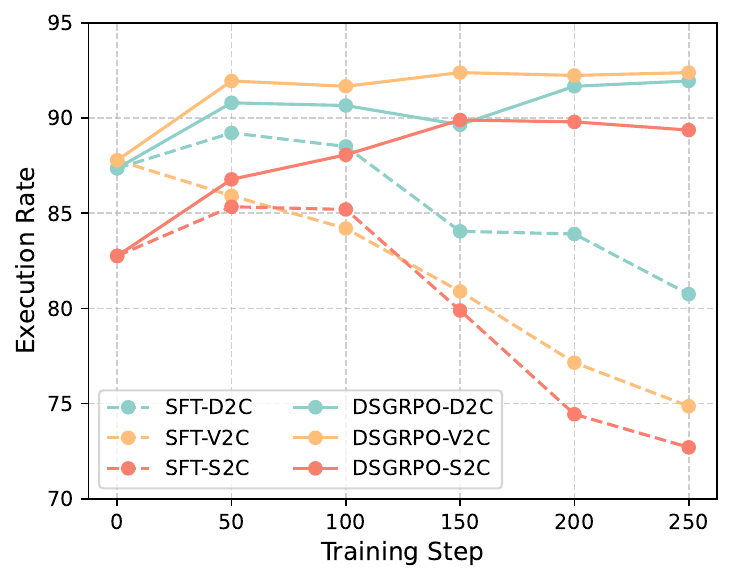}
    \end{minipage}
    \hfill
    \begin{minipage}[b]{0.31\textwidth}
        \centering
        \includegraphics[width=\textwidth]{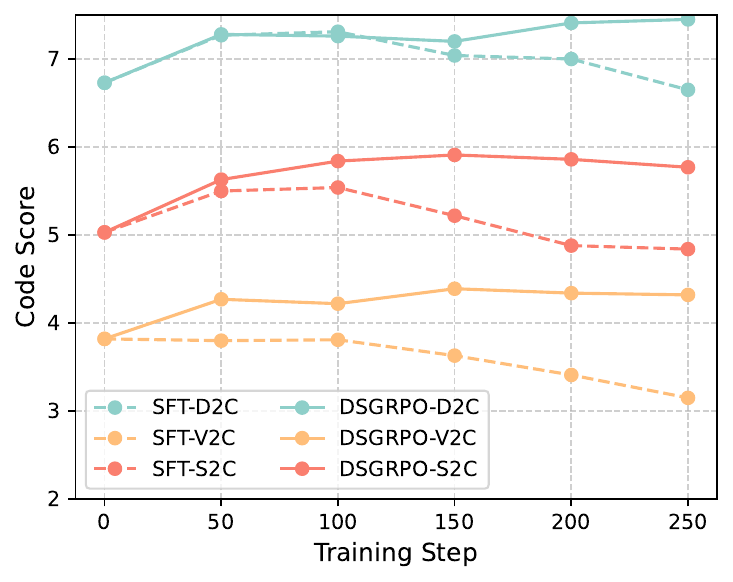}
    \end{minipage}
    \hfill
    \begin{minipage}[b]{0.31\textwidth}
        \centering
        \includegraphics[width=\textwidth]{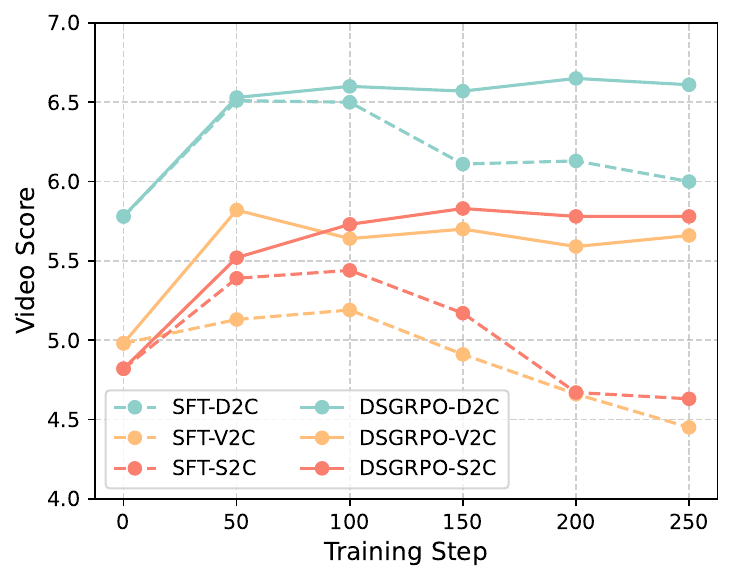}
    \end{minipage}
    \caption{Generalization Comparison Between SFT and JCV-GRPO on D2C, S2C, and V2C tasks across Execute Rate, Code Score, and Video Score metric.}
    \label{fig:generalize}
\end{figure*}

\subsection{Generalization of Joint-Code-Visual GRPO training}
To assess the generalization capabilities of JCV-GRPO, we compare JCV-GRPO with SFT based on Qwen2.5-DCG$_{s1}$. To ensure fair computational budget comparison, while using the same 1k D2C data to train Qwen2.5-DCG$_{s2}$, SFT training runs for 8 epochs to match the group size of 8 in JCV-GRPO. As shown in Figure~\ref{fig:generalize}, JCV-GRPO consistently improves execution rate, $S_{code}$, and $S_{video}$ throughout training, demonstrating generalization to S2C and V2C tasks. In contrast, continued supervised fine-tuning leads to overfitting, with significant performance degradation across all metrics compared to Qwen2.5-DCG$_{s1}$ baseline as training step growth. Ablation results demonstrate that JCV-GRPO training not only steadily improves model performance but also generalize well across tasks and modality to overcome memorization issue of long chart code.

\subsection{Ablation Study}
\begin{table}[h]
    \centering
    \small
    \setlength{\tabcolsep}{3pt}
    \begin{tabular}{c|c|ccc|ccc|ccc}
        \toprule
        \multicolumn{2}{c|}{Data} & \multicolumn{3}{c}{D2C} & \multicolumn{3}{c}{S2C} & \multicolumn{3}{c}{V2C} \\
        D2C & V2C & $Eexc.\ Rate$ & $S_{code}$ & $S_{video}$ & $Eexc.\ Rate$ & $S_{code}$ & $S_{video}$ & $Eexc.\ Rate$ & $S_{code}$ & $S_{video}$ \\
        \midrule
        \ding{51} & \ding{55} & \underline{80.17} & \textbf{6.53} & \underline{5.67} & 75.29 & \underline{4.51} & \underline{4.34} & 70.89 & 1.86 & 2.99 \\
        \ding{55} & \ding{51} & 69.68 & 4.11 & 4.12 & \underline{75.57} & 4.06 & 4.11 & \underline{82.90} & \underline{3.57} & \underline{4.90} \\
        \ding{51} & \ding{51} & \textbf{84.91} & \underline{6.33} & \textbf{5.69} & \textbf{81.47} & \textbf{4.95} & \textbf{4.88} & \textbf{85.06} & \textbf{3.61} & \textbf{4.98} \\
        \bottomrule
    \end{tabular}
    \caption{Ablation study of SFT training data composition, including mixture of D2C data and V2C data.}
    \label{tab:sft_data_ablation}
\end{table}

\begin{figure*}[htp]
    \centering
    \begin{minipage}[b]{0.31\textwidth}
        \centering
        \includegraphics[width=\textwidth]{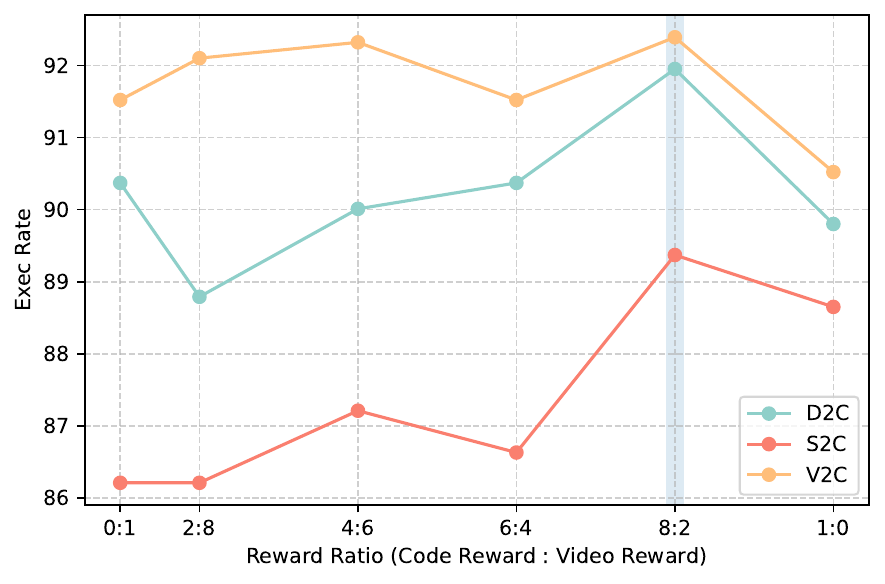}
    \end{minipage}
    \hfill
    \begin{minipage}[b]{0.31\textwidth}
        \centering
        \includegraphics[width=\textwidth]{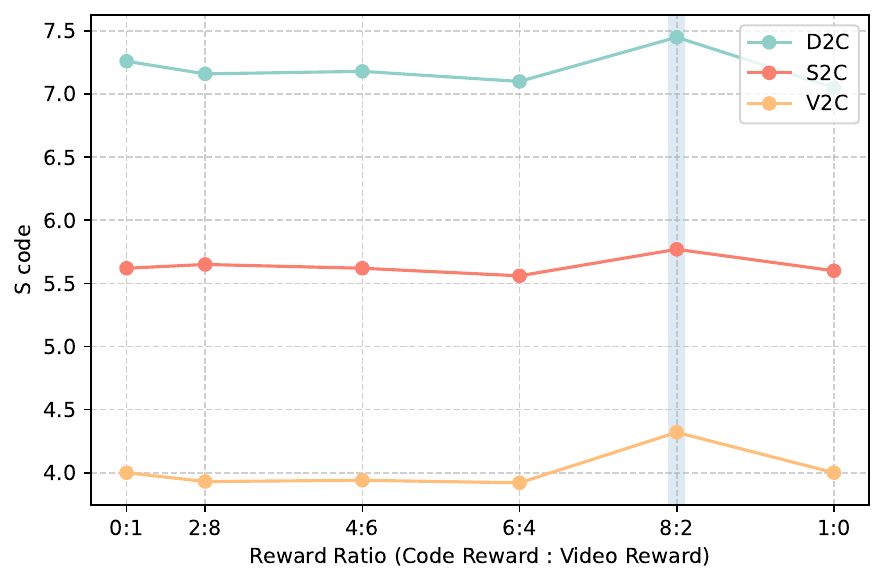}
    \end{minipage}
    \hfill
    \begin{minipage}[b]{0.31\textwidth}
        \centering
        \includegraphics[width=\textwidth]{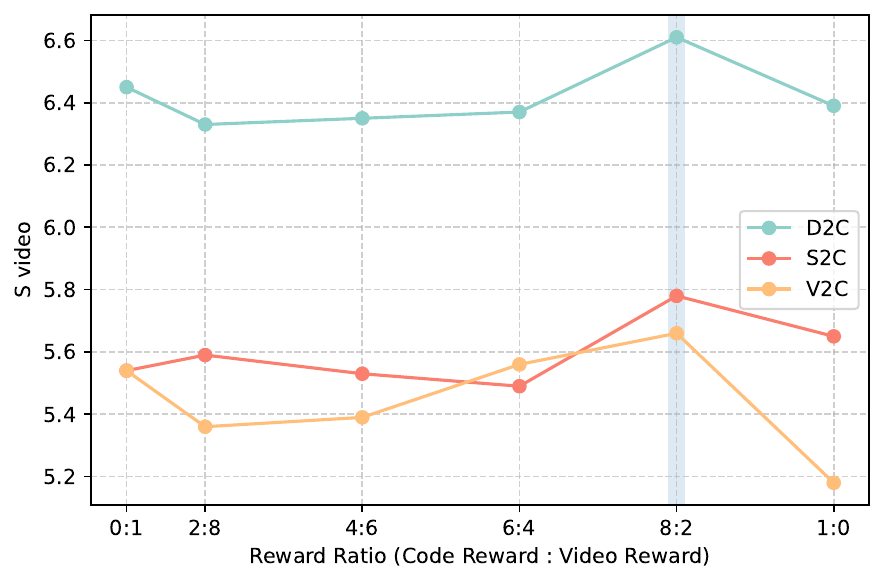}
    \end{minipage}
    \caption{Ablation study of reward score ratio in JCV-GRPO training, covering ratio from 1:0 to 0:1 for code:video reward}
    \label{fig:reward_ratio_ablation}
\end{figure*}

\textbf{SFT Data Composition.} 
We conducted an ablation study on SFT data composition. As shown in Table~\ref{tab:sft_data_ablation}, SFT on a single task can only achieve higher performance on the trained task, showing limited generalization. It is notable that although trained solely on the V2C task, the model can achieve competitive performance on the S2C task compared to models trained solely on the D2C task, demonstrating visual-to-language transfer potential in the dynamic chart generation task. Combining both modalities, with half the data from each task type, boosts performance on both tasks together, achieving the highest execution rate, $S_{code}$ and $S_{video}$ across all three tasks.

\noindent
\textbf{Reward Score Ratio.} Figure~\ref{fig:reward_ratio_ablation} presents ablation study on reward composition used in JCVR-GRPO training. Gradually increasing the weight of the code reward while maintaining a modest video-level reward leads to monotonic improvements in execution success rate, $S_{\text{code}}$, and $S_{\text{video}}$, with all three tasks reaching their peak performance at a ratio of $\text{Code}\!:\!\text{Video} = 8\!:\!2$.. Eliminating video feedback degrades performance, showing that certain visual signals serve as auxiliary to code reward, whereas allowing code reward to dominate sharply diminishes execution reliability.

\subsection{User Study and Qualitative Result}
\begin{figure}[h]
    \centering
    \includegraphics[width=\textwidth]{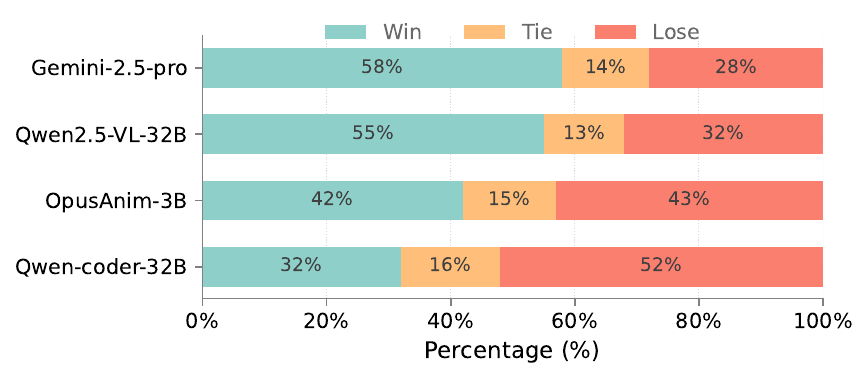}
    \caption{Human pairwise preference evaluation on Simple text to Code task.}
    \label{fig:user_study}
\end{figure}

\noindent
\textbf{User Study.} We conduct user study to compare the quality of the generated dynamic chart on V2C task. As shown in figure~\ref{fig:user_study}, Qwen2.5-DCG-3B obtained a 42\% win rate, outperforming the Qwen-2.5-coder-32B. The result highlights that Qwen2.5-DCG-3B shows strong capabilities but still demonstrates some gaps compared to proprietary models. We provide additional user studies on other tasks and detailed correlation analysis in the Appendix.

\begin{figure}[h]
    \centering
    \includegraphics[width=\textwidth]{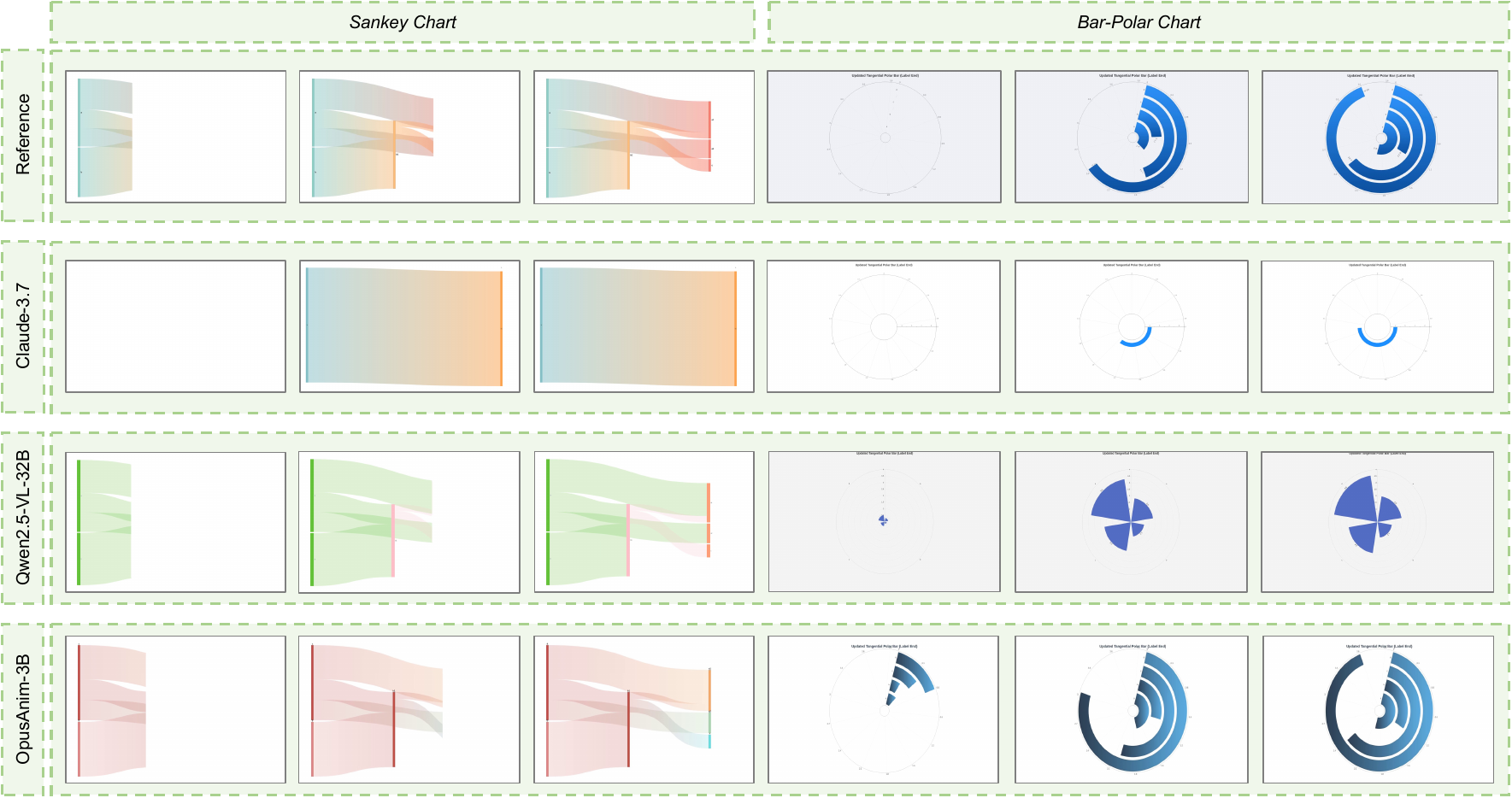}
    \caption{Qualitative Result Visualization.}
    \label{fig:qualitative_result}
\end{figure}

\noindent
\textbf{Qualitative Result.} Figure~\ref{fig:qualitative_result} presents representative generation results. Under the same descriptions, Qwen2.5-DCG-3B generates well-structured dynamic chart animations that more closely align with the reference videos compared to other open-source models. Additional examples and animation videos are provided in the supplementary material.

\subsection{Further Analysis}
\begin{figure}[htp]
    \centering
    \includegraphics[width=\textwidth]{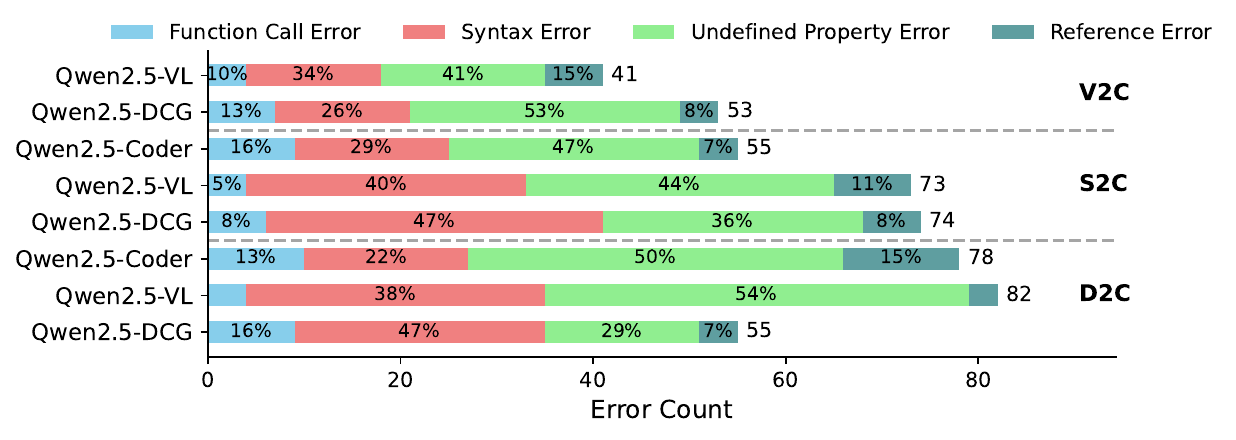}
    \caption{Error Analysis.}
    \label{fig:error_analysis}
\end{figure}
We further analyzed the types of code errors. Qwen2.5-DCG-3B$_{s2}$ exhibited lower rates of Undefined Property Errors on the S2C and D2C tasks, which are typically caused by incorrect code structure, indicating improved code generation in language-grounded tasks. On the V2C task, our model also achieved a 7\% lower syntax error rate compared to Qwen2.5-VL-32B. Additionally, Qwen2.5-Coder-32B recorded the lowest syntax error rates on D2C and S2C tasks, suggesting that adopting a CodeLLM backbone may yield a stronger DCG expert~\cite{zhao2025chartcoder}, which we leave for future work.

\section{Conclusion}
In this work, we introduce DCG-Bench, the first-ever benchmark for evaluating MLLMs in the dynamic chart generation task. Our benchmark analysis reveals limitations of existing MLLMs in the video-to-chart generation task, and explores a training recipe to construct expert MLLMs for the DCG task, achieving outstanding performance on all D2C, S2C, and V2C tasks, exceeding the best open-source MLLM Qwen2.5-VL-32B by 11.9\% with only 3B parameters. Furthermore, our training recipe reveals the high efficiency of knowledge transfer across input modalities, where SFT on V2C tasks brought significant improvement on D2C and S2C tasks as well, while GRPO training on D2C tasks led to V2C performance improvements. Our work provides a valuable resource and training insights for advancing research in dynamic chart generation and multimodal code generation areas.

\noindent \textbf{Limitations.}
While our approach shows strong performance in dynamic chart generation, several limitations remain. First, the current model architecture and RL optimization strategy, though effective, are not fully explored. Future work could investigate alternative designs and more advanced RL algorithms tailored to the DCG setting. Second, relying on Gemini as the reward model introduces dependence on a proprietary system. Open-sourced, QA-based reward mechanisms may offer better transparency and reproducibility.

\newpage

\bibliography{main}

\newpage
\appendix

\end{CJK*}
\end{document}